\documentclass[sn-basic,iicol]{sn-jnl}

\jyear{2023}%

\theoremstyle{thmstyleone}%
\usepackage{multirow}
\usepackage{colortbl}
\newcommand{\myPara}[1]{\vspace{12pt} \noindent\textbf{#1} }
\usepackage{bbding}
\usepackage[normalem]{ulem}
\useunder{\uline}{\ul}{}

\begin{document}

\title[Article Title]{Towards Training-free Open-world Segmentation via Image Prompt Foundation Models}

\author[1]{\fnm{Lv} \sur{Tang}}\email{lvtang@vivo.com}
\equalcont{These authors contributed equally to this work.}

\author[1]{\fnm{Peng-Tao} \sur{Jiang}}\email{pt.jiang@vivo.com}
\equalcont{These authors contributed equally to this work.}

\author[1]{\fnm{Haoke} \sur{Xiao}}\email{hk.xiao.me@gmail.com}
\equalcont{These authors contributed equally to this work.}

\author*[1]{\fnm{Bo} \sur{Li}}\email{libra@vivo.com}

\affil[1]{\orgdiv{vivo Mobile Communication Co., Ltd}, \orgaddress{\city{Shanghai}, \country{China}}}


\abstract{
The realm of computer vision has witnessed a paradigm shift with the advent of foundational models, mirroring the transformative influence of large language models in the domain of natural language processing. This paper delves into the exploration of open-world segmentation, presenting a novel approach called Image Prompt Segmentation (IPSeg) that harnesses the power of vision foundational models. 
IPSeg lies the principle of a training-free paradigm, which capitalizes on image prompt techniques. Specifically, IPSeg utilizes a single image containing a subjective visual concept as a flexible prompt to query vision foundation models like DINOv2 and Stable Diffusion. Our approach extracts robust features for the prompt image and input image, then matches the input representations to the prompt representations via a novel feature interaction module to generate point prompts highlighting target objects in the input image. The generated point prompts are further utilized to guide the Segment Anything Model to segment the target object in the input image. 
The proposed method stands out by eliminating the need for exhaustive training sessions, thereby offering a more efficient and scalable solution.
Experiments on COCO, PASCAL VOC, and other datasets demonstrate IPSeg's efficacy for flexible open-world segmentation using intuitive image prompts. This work pioneers tapping foundation models for open-world understanding through visual concepts conveyed in images.
}


\maketitle

\section{Introduction}
In recent years, large language models (LLMs)~\citep{chowdhery2023palm,touvron2023llama,zhang2022opt} have sparked a revolution in natural language processing (NLP). These foundational models exhibit remarkable transfer capabilities, extending far beyond their initial training objectives. LLMs showcase robust generalization abilities and excel in a multitude of open-world language tasks, including language comprehension, generation, interaction, and reasoning. Inspired by the success of LLMs, vision foundational models such as CLIP~\citep{DBLP:conf/icml/RadfordKHRGASAM21}, DINOv2~\citep{DBLP:journals/corr/abs-2304-07193}, BLIP~\citep{DBLP:conf/icml/0001LXH22}, and SAM~\citep{kirillov2023segment} have also emerged. These models, once trained, can seamlessly apply their knowledge to various downstream tasks.
Such a trend has further motivated researchers to explore ways of open-world visual understanding.

\begin{figure*}[!htbp]
    \centering
    \includegraphics[width=\linewidth]{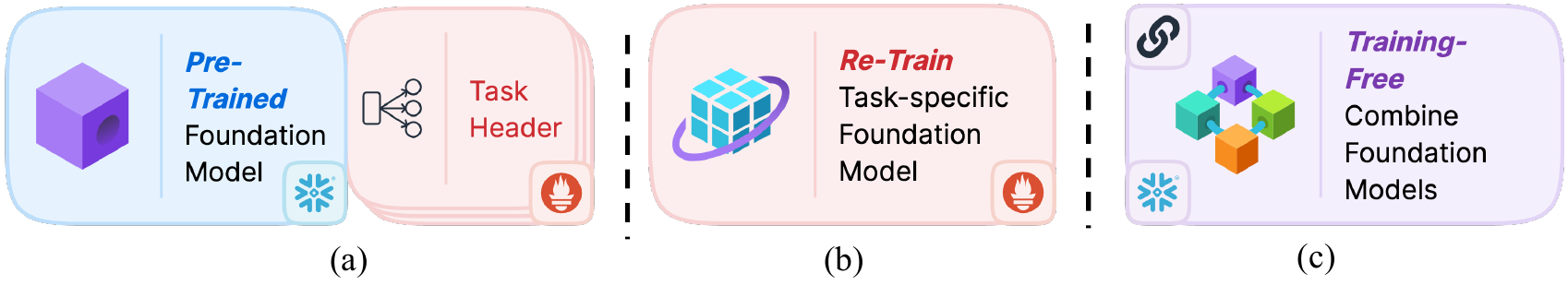}
    \caption{{Comparison of different open-world segmentation frameworks based on foundation models. From left to right, they are foundation model adaptions, task-specific foundation models training from scratch, and training-free foundation models. }}
    \label{introd1}
\end{figure*}

Pioneering works~\citep{DBLP:journals/corr/abs-2304-08485,DBLP:journals/corr/abs-2305-06500,DBLP:journals/corr/abs-2304-10592} have mainly focused on how to understand images as a whole in the open world. Herein, we project our viewpoint to open-world understanding at the object level, specifically for the task of open-world segmentation \citep{qi2022open}. 
When approaching open-world segmentation tasks, there are three primary strategies for leveraging foundational models. 
The most widely studied approach~\citep{DBLP:conf/cvpr/LiangWDLZ0ZVM23,DBLP:conf/cvpr/OinWYLRXWWWPW23,DBLP:conf/eccv/GhiasiGCL22} is to utilize a vision foundation model like CLIP or DINOv2 and cooperate it with a specific segmentation header or adapter to complete the open-world segmentation task. Such methods (Fig. \ref{introd1}a) often require fine-tuning or training the segmentation header or adapter. 
In addition to the above methods combining foundation model with adapter, some researchers have tried to draw on the successful experience in NLP and directly train a foundation model for generic dense-prediction vision problems, as demonstrated in works like Painter~\citep{DBLP:conf/cvpr/WangWCS023}. Such models (Fig. \ref{introd1}b) can complete open-world segmentation simply with a task-specific prompt. 
Lately, the Segment Anything Model (SAM)~\citep{kirillov2023segment} has attained remarkable zero-shot segmentation results. 
It presents researchers with the prospect of devising an alternative way to accomplish open-world segmentation without the need for training (Fig. \ref{introd1}c). For example, PerSAM~\citep{zhang2023personalize} effectively transfers SAM to open-world object segmentation tasks in a training-free manner through the design of the cross-attention layer in SAM's decoder, thereby tapping into the potential of vision foundational models to a significant extent. 
While these approaches have achieved excellent performance, incorporating more vision foundation models to improve the generalization capability and segmentation quality for open-world segmentation remains an avenue for further inspection.

\begin{figure*}[!htbp]
    \centering
    \includegraphics[width=\linewidth]{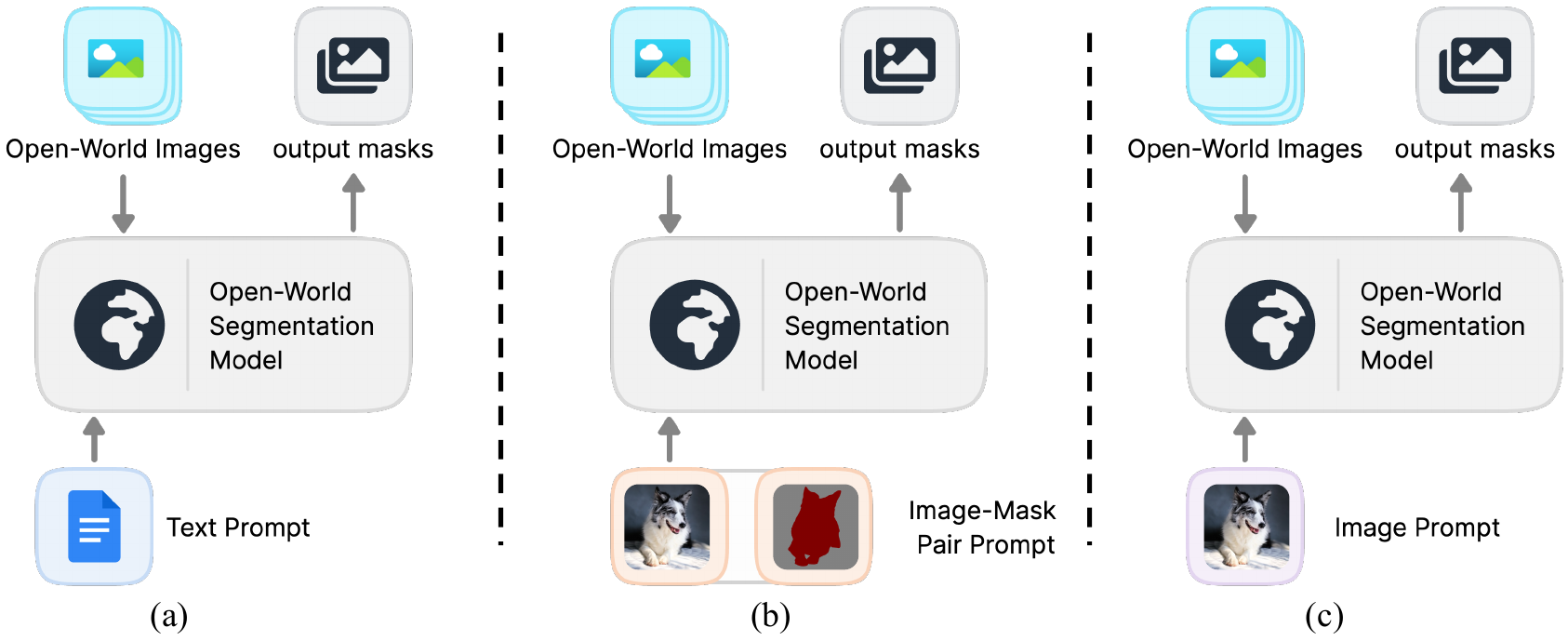}
    \caption{{ Different prompt forms in existing open-world segmentation methods. The left is the prompt of predefined 
    textual descriptions or categories. The middle is the prompt form used in existing one-shot object segmentation works \citep{liu2023matcher,zhang2023personalize}. The right is the prompt form used in this paper, which only uses one image containing a salient object with specific visual concepts.}}
    \label{introd2}
\end{figure*}

In addition to the architectural design of the foundation model for open-world segmentation tasks, another critical aspect is the development of  flexible and user-friendly prompts. This ensures that the model accurately grasps the visual concepts users desire.
As shown in Fig. \ref{introd2}(a)(b), existing works typically rely on predefined textual descriptions or high-quality annotations for a given image as the segmentation prompt, which lacks flexibility. Yet, in the context of open-world scenarios, we not only expect the network to perform well on various open-set datasets but also need it to handle object segmentation tasks with more versatile prompt information. Therefore, a fundamental question emerges: Could we prompt the foundational models, such as SAM, to segment specific objects based on the prompt of the user-given image that contains objects with a clear subjective concept?

Motivated by this question, we present a novel open-world segmentation framework, which utilizes image prompts to instruct the training-free vision foundational models to segment open-world objects.
The proposed Image Prompt Segmentation (IPSeg) network is a straightforward yet highly effective framework, comprising three main components, \emph{i.e.}, feature extraction, 
feature interaction, and segmentation. 
For the feature extraction, we design two branches, including the \textbf{prompt} and 
the \textbf{input} branches.
The \textbf{prompt} branch is dedicated to capturing general representations of subjective objects belonging to a specific category from the prompt image, and the extracted representations are employed to identify the objects in the input image. 

The \textbf{input} branch is designed to capture the feature representation of the input image to be segmented, following the same architecture proposed in the \textbf{prompt} branch.
For the feature interaction, we've devised a feature interaction module to facilitate interaction between the input image features and the given image prompt features, thereby accentuating the pixel points of the target objects. 
Finally, the generated pixel points serve as the prompt information for SAM \citep{kirillov2023segment}, guiding SAM in predicting the final segmentation map. 

In summary, the key contributions are listed as follows:

\begin{itemize}
\item We propose a training-free open-world object segmentation framework based on foundational models. We take the pioneering step of utilizing image prompts with clear target objects to query generic object representations from foundational 
models. Such a framework can potentially inspire researchers to address open-world segmentation from a fresh perspective. 

\item We introduce a simple but effective framework, coined as IPSeg, which contains three effective components. They are utilized to extract discriminative features of target objects identified in the given image prompt and generate accurate points to prompt SAM models to generate object masks. 

\item We validate the proposed IPSeg framework on widely used segmentation datasets, including COCO-20$^i$~\citep{DBLP:conf/iccv/NguyenT19}, FSS-1000~\citep{DBLP:conf/cvpr/LiWCTT20} and PerSeg~\citep{zhang2023personalize}. Compared to methods PerSAM and Painter, our proposed method can achieve a 30.6\% and 42.8\% improvement in the mIoU metric with flexible prompts under a training-free mechanism. \end{itemize}

\section{Related works}

\subsection{Large Vision Models (LVMs)}
Prompted by the powerful generalized ability of large language models \citep{devlin2018bert,lu2019vilbert,brown2020language,radford2018improving,radford2019language,zhang2023llama} 
in nature language processing, large vision models \citep{DBLP:journals/corr/abs-2304-07193, kirillov2023segment,DBLP:conf/icml/RadfordKHRGASAM21} have emerged.
Among these large vision models, CLIP \citep{DBLP:conf/icml/RadfordKHRGASAM21} align the image and text feature spaces through contrastive learning on the huge number of image-text pairs, whose models show powerful zero-shot generalization ability on various downstream vision tasks \citep{xu2023side}, 
such as open-world segmentation \citep{qi2022open,cen2021deep}.
SAM \citep{kirillov2023segment} train a prompt-based large 
segmentation model on 1 billion masks.
The prompt-based segmentation model can accurately segment objects 
in images from different domains.
Such ability has facilitated different applications, 
such as object tracking \citep{yang2023track,cheng2023segment,zhu2023tracking}, image segmentation \citep{zhang2023customized,chen2023segment,tang2023can,jiang2023segment}, 3D reconstruction \citep{cen2023segment,shen2023anything} etc.
Besides, DINOv2 \citep{DBLP:journals/corr/abs-2304-07193} learn powerful 
object-level representations in an unsupervised manner.
Such powerful representations facilitate downstream dense 
scene parsing tasks, such as semantic segmentation \citep{chen2017deeplab,long2015fully}, 
and depth estimation \citep{ranftl2021vision}.

\subsection{Open-World Segmentation}
Open-world segmentation aims to extend traditional close-set segmentation models \citep{long2015fully,chen2017deeplab}
to enable open-set pixel classification, making them more versatile and capable of generalization.
The models of open-world segmentation \citep{cui2020open,cen2021deep,qi2022open} need to be able to handle unknown classes.
There exist several kinds of open-world segmentation methods.
The first line of works attempts \citep{xia2020synthesize,cen2021deep,angus2019efficacy,hammam2023identifying} to classify the pixels of objects out of the training set's distribution to `anomaly'.
They do not distinguish different novel classes in ``anomaly", in detail.
The second line of works \citep{xian2019semantic,bucher2019zero} usually trains segmentation models on datasets with a fixed number of seen classes and utilizes the models to segment images with unseen classes.
They strive to improve the generalization of segmentation embedding to unseen classes.

\begin{figure*}[]
    \centering
    \includegraphics[width=0.85\linewidth]{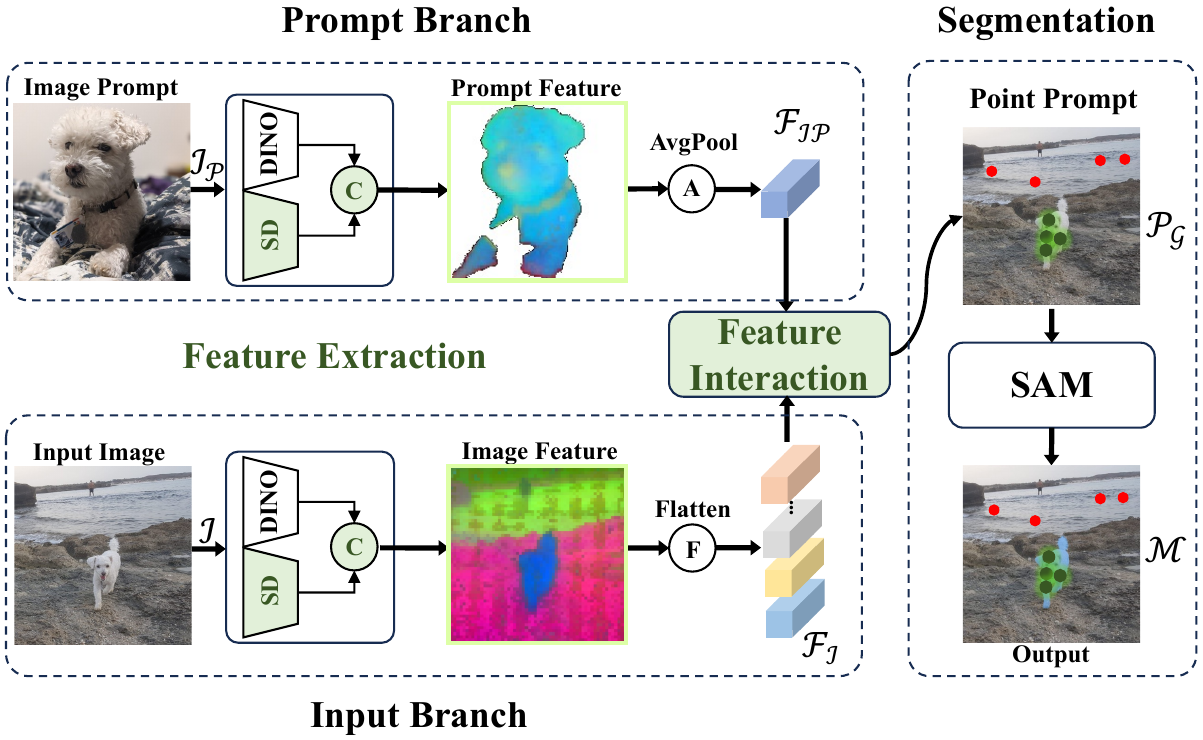}
    \caption{The framework of our proposed IPSeg framework. Importantly, all parameters in the network remain frozen, eliminating the need for additional training. The {\color{green} green} point in $\mathcal{P}_\mathcal{G}$ represents the positive point prompts sent to SAM, while the {\color{red}red} point represents the negative point prompts sent to SAM.}
    \label{framework}
\end{figure*}

Recently, owing to LVMs, such as CLIP \citep{DBLP:conf/icml/RadfordKHRGASAM21} have shown significant zero-shot 
classification ability, researchers attempt to transfer their image-level classification ability to region-level classification.
These methods \citep{luo2023segclip,xu2023side,ma2022open,DBLP:conf/cvpr/LiangWDLZ0ZVM23,xu2022simple,zhou2023zegclip,liu2022open} adapt CLIP models to the open-world segmentation models by training on the datasets with seen classes to align the predicted region features and text features.
Among the methods using LVMs, some works \citep{zhou2022extract,liu2023matcher,zhang2023personalize} also attempt to utilize training-free LVMs and design prompts to conduct open-world segmentation.
Without fine-tuning LVMs, they directly extract object segmentation masks from them. 
Zhou et al. \citep{zhou2022extract} conduct minimal modification of the CLIP model to extract segmentation masks of open-world categories.
Liu et al. \citep{liu2023matcher} and Zhang et al. \citep{zhang2023personalize} utilize an image 
with an object mask to extract prompts.
Then, the prompts are used to instruct 
the SAM model to segment 
objects of the target category indicated in the provided image.

Our proposed method also falls into the training-free LVMs categories.
Different from previous works using image-mask pairs, we only utilize an image containing the objects of the target concept as prompts to conduct open-world segmentation.
Image prompts are more flexible than image-mask pairs, as 
humans do not need to annotate the objects of the target class.
Besides, we also utilize off-of-the-shelf LVMs, such as DINOv2, to extract discriminative feature representations of image prompts.
Then, discriminative feature representations are used to prompt LVMs to segment target objects in test images.


\section{Method}
We first introduce the preliminaries about the Segmentation Anything Models (SAM) \citep{kirillov2023segment}, used in this paper. Then, we introduce the proposed IPSeg framework, which is shown in Fig. \ref{framework}. Given an image prompt with a clear concept, IPSeg is capable of segmenting any semantically identical object under the open-world setting.

\subsection{Preliminaries}
SAM consists of three components: a prompt encoder Enc$\mathcal{P}$, an image encoder Enc$\mathcal{I}$, and a lightweight mask decoder Dec$\mathcal{M}$. As a prompt-based framework, SAM takes as input an image $\mathcal{I}$, and prompts $\mathcal{P}$ (like specific points). Specifically, SAM initially utilizes Enc$\mathcal{I}$ to extract features from the input image and employs Enc$\mathcal{P}$ to encode the provided prompts into prompt tokens:
\begin{align}
    F_\mathcal{I} = Enc\mathcal{I}(\mathcal{I}), \quad T_\mathcal{P} = Enc\mathcal{P}(\mathcal{P}).
\end{align}
Afterwards, the encoded image $F_\mathcal{I}$ and prompts $T_\mathcal{P}$ are input into the decoder Dec$\mathcal{M}$ for feature interaction. It's worth noting that SAM constructs the decoder's input by concatenating several learnable mask tokens $T_\mathcal{M}$ as prefixes to the prompt tokens $T_\mathcal{P}$. These mask tokens are responsible for generating the mask output, formulated as:
\begin{align}
    \mathcal{M} = Dec\mathcal{M}(F_\mathcal{I}, Concat(T_\mathcal{M},T_\mathcal{P})),
\end{align}
where $\mathcal{M}$ denotes the final segmentation masks predicted by SAM.

As discussed above, SAM can segment objects in an image based on the given prompt. Therefore, the core of this paper lies in how to find semantically matching points in the image $\mathcal{I}$ to be segmented when given an image prompt $\mathcal{I}_{\mathcal{P}}$ that contains clear visual concepts. This, in turn, guides SAM in generating segmentation results.
Note we focus on constructing an image-prompt open-world framework.
Exploring prompts, like bounding boxes, is out of the scope of this paper.

\subsection{Overview}
The pipeline of our method is shown in Fig. \ref{framework}.
The proposed IPSeg framework comprises three components: feature extraction, feature interaction and SAM. The feature extraction module is used in the \textbf{prompt} branch and \textbf{input} branch, which can extract the discriminative feature representations of both input image $\mathcal{I}$ and image prompt $\mathcal{I}_{\mathcal{P}}$. Then, the prompt feature $\mathcal{F}_\mathcal{IP}$ interacts with the input image feature $\mathcal{F}_\mathcal{I}$ in the feature interaction module, to generate specialized prompts $\mathcal{P}_\mathcal{G}$ such as points in the input image, which contains the same semantic information with the prompt image. Finally, the generated prompt $\mathcal{P}_\mathcal{G}$ and the input image $\mathcal{I}$ are sent to SAM, generating the final prediction $\mathcal{M}$. We will provide detailed explanations of the first two components in the subsequent subsections.

\begin{figure*}[!htbp]
    \centering
    \includegraphics[width=0.95\linewidth]{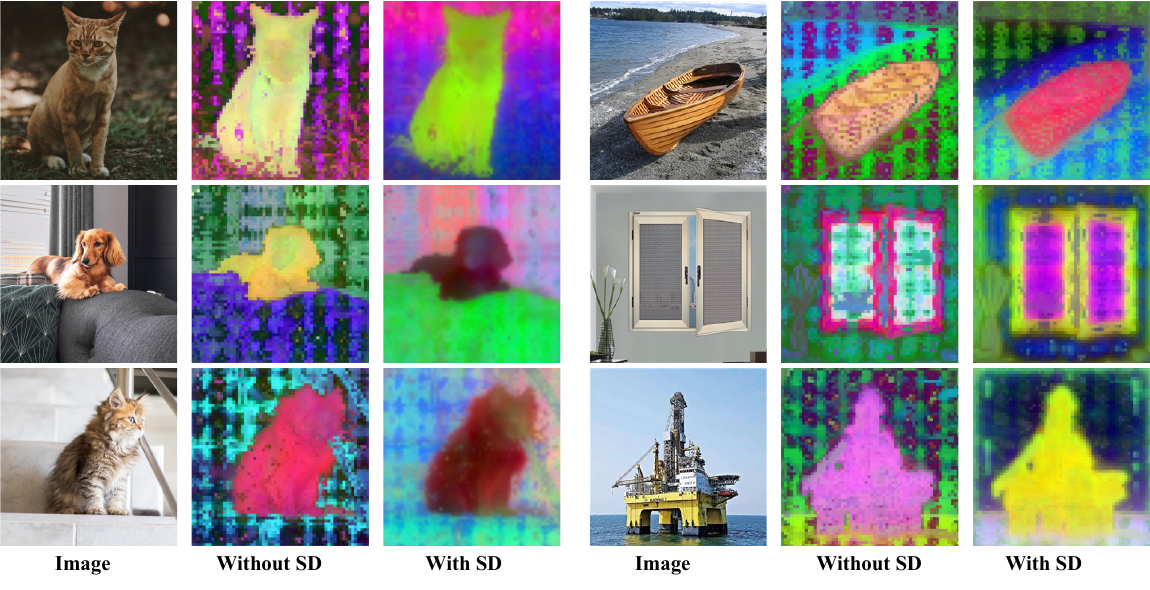}
    \caption{Visualization results of features extracted from different models.
    The second and fifth columns indicate the use of only the DINOv2 model for feature extraction, while the third and sixth columns denote the use of both DINOv2 and SD models for this purpose.}
    \label{dinosd}
\end{figure*}

\subsection{Feature Extraction}
Extracting a robust feature representation from both the prompt image $\mathcal{I}_{\mathcal{P}}$ and the input image $\mathcal{I}$, which effectively captures the visual semantic information in both sets of images, also ensures that the network can find a consistent semantic object between these two sets of images. Generally, the feature representation of an image can be divided into high-level feature representation and low-level feature representation. In this paper, we explore how to extract a feature representation of an image from both of these aspects.

In the following, we first introduce the feature extraction process. Then, we introduce how we utilize the feature extraction to constitute the \textbf{prompt} and \textbf{input} branch of the IPSeg framework.

\subsubsection{Feature Extraction}

\myPara{High-level Feature Extraction}
Previous study \citep{DBLP:journals/corr/abs-2304-07193} has established that features from Vision Transformers, particularly those from DINOv2, are rich in explicit information pertinent to semantic segmentation and are highly effective when used as K-Nearest Neighbors classifiers. DINOv2, in essence, excels at extracting semantic content with high accuracy from each image. Consequently, we have chosen to utilize the features extracted by the foundational model DINOv2 to represent the semantic information of each image, denoted as $\mathcal{F}_\mathcal{D}$.

\begin{figure*}[!htbp]
    \centering
    \includegraphics[width=\linewidth]{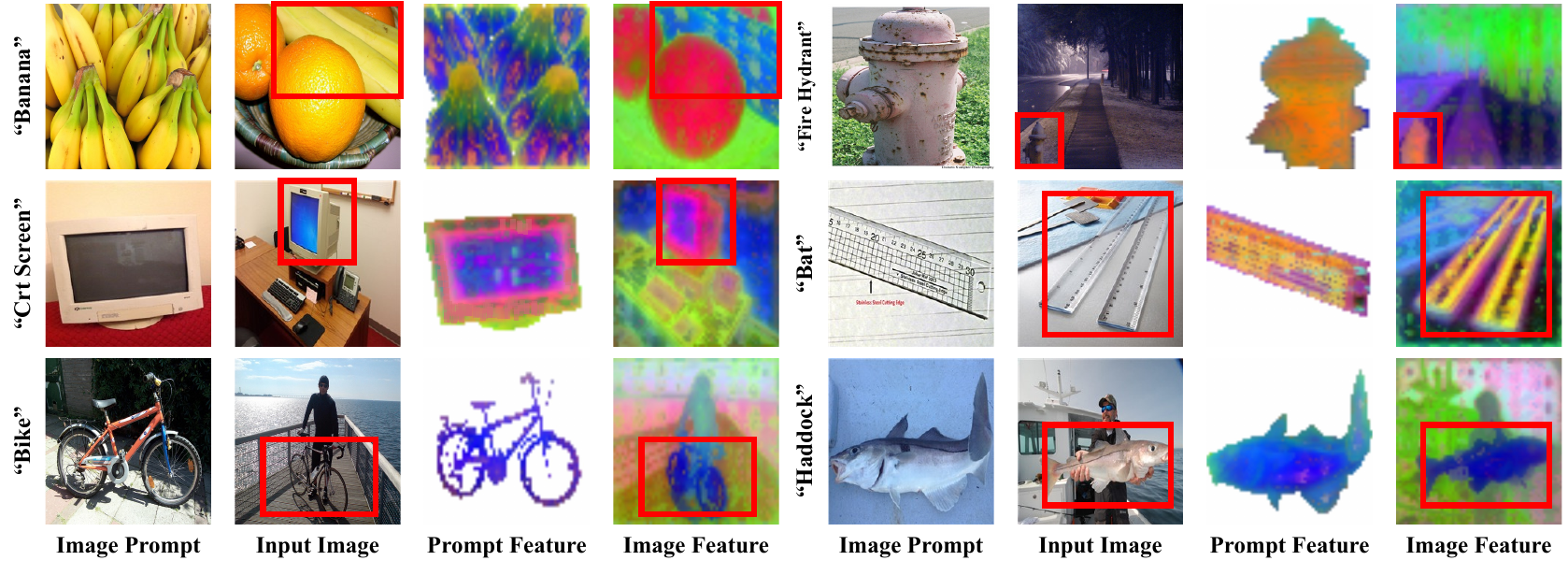}
    \caption{Visualizing the features of foreground objects in the prompt image and all objects in input prompt.}
    \label{promptfeature}
\end{figure*}

\myPara{Low-level Feature Extraction}
DINOv2 is proficient in capturing significant high-level semantic information, yet it has limitations in providing intricate low-level detail information. As illustrated in the second column of Fig. \ref{dinosd}, the visual features generated exclusively through DINOv2 might miss out on fine-grained low-level details. Notably, there is a discernible research gap in augmenting features extracted by DINOv2 with low-level detail information without necessitating additional training.

In our proposed IPSeg, integrating a pre-trained model that specializes in capturing low-level detail information becomes vital. Such a model is capable of effectively compensating for the detailed information that might be overlooked by DINOv2. Notably, Stable Diffusion (SD) \citep{DBLP:conf/cvpr/RombachBLEO22} has recently been recognized for its exceptional prowess in generating high-quality images, underscoring its ability to robustly represent images with comprehensive content and detailed information. Consequently, our primary focus is to explore the potential benefits of combining SD features with DINOv2 in enhancing the overall quality of feature representations.

The architecture of SD consists of three key components: an encoder $\mathcal{E}_{nc}$, a decoder $\mathcal{D}_{ec}$, and a denoising U-Net $\mathcal{U}_{net}$ operating within the latent space. We initiate the process by projecting an input image $I_0$ into the latent space using the encoder $\mathcal{E}_{nc}$, resulting in a latent code $x_0=\mathcal{E}_{nc}(I_0)$. Subsequently, we introduce Gaussian noise $\epsilon$ to the latent code, following a predefined time step $t$. Finally, utilizing the latent code $x_t$ at time step $t$, we extract the SD features $\mathcal{F}_\mathcal{S}$ through the denoising U-Net:
\begin{align}
    \mathcal{F}_\mathcal{S} = \mathcal{U}_{net}(x_t,t),\ x_t = \sqrt{\Bar{a}_t}x_0 + \sqrt{1-\Bar{a}_t}\epsilon.
\end{align}
$\Bar{a}_t$ is utilized to determine the noise schedule~\citep{ho2020denoising}.

\myPara{Feature Fusion}
Building upon the discussions mentioned earlier, we present a straightforward yet notably effective fusion strategy. This strategy is designed to capitalize on the strengths of both SD and DINOv2 features:
\begin{align}
    \mathcal{F}_\mathcal{F} =  Cat(\mathcal{F}_\mathcal{S},
 \mathcal{F}_\mathcal{D}),
 \label{fuse}
\end{align}
where $Cat(,)$ denotes 
feature concatenation along the channel dimension. In the third and sixth columns of Fig. \ref{dinosd}, the fused feature aids in generating a smoother and more resilient visual feature, which helps for feature matching. Specifically, the addition of SD enhances the internal features of foreground objects, making them smoother and more consistent, thereby assisting the network in extracting target objects from segmented images.

\subsubsection{Input and Prompt Branches }
After introducing the pipeline of the feature extraction, we utilize the visual 
encoder to extract features for the input image (\textbf{input} branch) and image prompt (\textbf{prompt} branch), respectively.

\myPara{Input Branch} 
For the input image $\mathcal{I}$, we use the above process to extract the feature $\mathcal{F}_\mathcal{I} \in \mathbb{R}^{H \times W \times C}$, where $H, W$ mean the spatial resolution of the feature and $C$ means the channel number. Then, we reshape the $\mathcal{F}_\mathcal{I}$ to $\mathbb{R}^{HW \times C}$, where $HW$ means the total number in the feature and the representation of each pixel is $\mathbb{R}^{C}$.

\myPara{Prompt Branch} 
For the image prompt $\mathcal{I}_\mathcal{P}$, we also extract its feature $\mathcal{F}_\mathcal{IP} \in \mathbb{R}^{H \times W \times C}$ through the above process. Since we do not care about the background information of this feature, we use an unsupervised salient object detection method TSDN~\citep{DBLP:conf/cvpr/Zhou0YLX23} to filter these pixels belong to the background, then use the average pooling (\emph{Avgpool}) operation to generate the prompt embedding:
\begin{align}
    \mathcal{F}_{IP} = Avgpool(\mathcal{F}_{IP} \odot \mathcal{MS}),
    \label{avg pool}
\end{align}
where $\odot$ denotes pixel-wise multiplication. The object map $\mathcal{MS}$ is directly obtained by the unsupervised method TSDN.

\begin{table*}[]
\centering
\caption{{  Comparison of the few-shot semantic segmentation performance between our proposed method and five typical generalist models. Painter~\citep{DBLP:conf/cvpr/WangWCS023}, SegGPT~\citep{DBLP:conf/iccv/WangZCWS023} and DeLVM~\citep{guo2024dataefficient} are three methods which require the extra training process. PerSAM~\citep{zhang2023personalize} and Matcher~\citep{DBLP:journals/corr/abs-2303-02153} are two training-free few-shot works. Matcher-Z means the performance of Matcher under zero-shot setting. \textbf{Note that, for IPSeg, we do not utilize the ground truth corresponding to the prompt image to select its foreground object.} We report mIoU (\%) in this table.}}
\begin{tabular}{@{}c|c|ccccc|c|c@{}}
\toprule
        &           & \multicolumn{5}{c|}{\textbf{COCO-20$^i$}}               &                             &      \\ \cmidrule(lr){3-7}
\multirow{-2}{*}{\textbf{Methods}} &
  \multirow{-2}{*}{\textbf{Pub. \& Year}} &
  \textbf{Fold0} &
  \textbf{Fold1} &
  \textbf{Fold2} &
  \textbf{Fold3} &
  \textbf{Mean} &
  \multirow{-2}{*}{\textbf{FSS}} &
  \multirow{-2}{*}{\textbf{PerSeg}} \\ \midrule
Painter & CVPR 2023  & {\color[HTML]{333333} 31.2} & 35.3 & 33.5 & 32.4 & 33.1 & {\color[HTML]{333333} 61.7} & 56.4 \\
SegGPT  & ICCV 2023  & {\color[HTML]{333333} 56.3} & 57.4 & 58.9 & 51.7 & 56.1 & {\color[HTML]{333333} 85.6} & 95.5 \\
DeLVM &
  ArXiv 2024 &
  \cellcolor[HTML]{FFFFFF}12.6 &
  \cellcolor[HTML]{FFFFFF}13.6 &
  \cellcolor[HTML]{FFFFFF}10.1 &
  \cellcolor[HTML]{FFFFFF}10.5 &
  \cellcolor[HTML]{FFFFFF}{\color[HTML]{333333} 11.7} &
  36.9 &
  9.8 \\ \midrule
PerSAM  & ICLR 2024  & 23.1                        & 23.6 & 22.0 & 23.4 & 23.0 & 81.6                        & 89.5 \\
Matcher & ICLR 2024 & {\color[HTML]{333333} 52.7} & 53.5 & 52.6 & 52.1 & 52.7 & {\color[HTML]{333333} 87.0} & 94.9 \\ \midrule
Matcher-Z &
  ICLR 2024 &
  \cellcolor[HTML]{FFFFFF}{\color[HTML]{333333} 22.9} &
  \cellcolor[HTML]{FFFFFF}23.2 &
  \cellcolor[HTML]{FFFFFF}22.2 &
  \cellcolor[HTML]{FFFFFF}22.8 &
  \cellcolor[HTML]{FFFFFF}{\color[HTML]{333333} 22.8} &
  {\color[HTML]{333333} 81.2} &
  88.3 \\
Ours &
  Year 2024 &
  {\color[HTML]{333333} \textbf{40.9}} &
  \textbf{44.9} &
  \textbf{40.1} &
  \textbf{46.2} &
  \textbf{43.0} &
  {\color[HTML]{333333} \textbf{82.7}} &
  \textbf{92.7} \\ \bottomrule
\end{tabular}
\label{generalist}
\end{table*}

\subsection{Feature Interaction and Segment}
After generating input image feature $\mathcal{F}_\mathcal{I}$ and input prompt feature vector $\mathcal{F}_\mathcal{IP}$, we can obtain specific point prompt for the input image $\mathcal{I}$ by performing interaction between $\mathcal{F}_\mathcal{I}$ and $\mathcal{F}_\mathcal{IP}$.

Concretely, for input image feature which contains $HW$ pixels, the feature representation of each pixel is denoted as $\mathcal{F}_\mathcal{I}^{l}$, where $l \in [1,HW]$. Firstly, we calculate the correlation score between $\mathcal{F}_\mathcal{IP}$ and $\mathcal{F}_\mathcal{I}^{l}$ through cosine similarity. Secondly, we utilize a TopK algorithm to select the points in the input image that are most semantically similar to the prompt image, which are 
at position $P_{coord}$:
\begin{align}
     S = \mathcal{F}_{IP} \otimes \mathcal{F}_{I}, P_{coord} = {\text{TopK}}(S) \in \mathbb{R}^K,
\end{align}
where $\otimes$ means matrix multiplication. As shown in Fig. \ref{promptfeature}, The foreground object in the prompt image and the object to be segmented in the input image maintain good semantic consistency, ensuring the effectiveness of our TopK algorithm.

Finally, we further refine the $P_{coord}$ into $c$ clustering centers as the positive point prompts for SAM. In addition, using the same pipeline, we also selected $K$ points that are the least similar to the prompt image feature and clustered them into $c$ cluster centers as negative point prompts for SAM. 
We set $K=32$ and $c=4$ in this paper.  
The generated positive/negative point prompts and the input image $\mathcal{F}_\mathcal{I}$ are sent to SAM to predict final segmentation results $\mathcal{M}$. 

\section{Experiment}
\subsection{Experimental Setup}
We employ the Stable Diffusion v1.5 and DINOv2 models as our feature extractors.
The DDIM timestep in the denoising process is set to 50 by default. 
All experiments are conducted on a single RTX A6000 GPU with only 13G GPU memory. This means that our proposed training-free framework can run on cheaper graphics cards such as RTX3090, providing a good perspective for researchers with limited computing power to explore foundational models.

\begin{table*}[]
\centering
\caption{Comparison of the zero-shot segmentation performance between our proposed methods and seven typical specialist models. We report mIoU (\%) in this table. \textbf{NT}: Need Training.}
\begin{tabular}{@{}
>{\columncolor[HTML]{FFFFFF}}c |
>{\columncolor[HTML]{FFFFFF}}c |
>{\columncolor[HTML]{FFFFFF}}c |
>{\columncolor[HTML]{FFFFFF}}c |
>{\columncolor[HTML]{FFFFFF}}c |
>{\columncolor[HTML]{FFFFFF}}c @{}}
\toprule
\textbf{Methods} & \textbf{Pub. \&Year} & \textbf{NT} & \textbf{COCO-Stuff}     & \textbf{PASCAL-VOC} & \textbf{PASCAL-Context} \\ \midrule
SPNet            & CVPR 2019            &  \checkmark   & 8.7                    & 15.6               & 4.0                    \\
ZS3              & NeurIPS 2019         &  \checkmark   & 9.5                    & 17.7               & 7.7                    \\
CaGNet           & MM 2020           &  \checkmark   & 13.9                   & 29.9               & 15.0                   \\
SIGN             & ICCV 2021            &  \checkmark   & 15.5                   & 28.9               & 14.9                   \\
ViL-Seg          & ECCV 2022            &  \checkmark   & 16.4                   & 34.4               & 16.3                   \\
GroupVit         & CVPR 2022            & \checkmark    & 16.1                   & 79.0               & 49.2                   \\
TCL          & CVPR 2023                & \checkmark    & 27.6                   & 84.5               & 62.0                   \\ \hline
Ours             & Year 2024            & \XSolidBrush  & \textbf{32.7} & \textbf{57.9}                    &  \textbf{67.7}                       \\ \bottomrule
\end{tabular}
\label{Specialist}
\end{table*}

\subsection{Evaluation Datasets}
Following PerSAM~\citep{zhang2023personalize}, we conduct few-shot experiments on three datasets, including COCO-20$^i$~\citep{DBLP:conf/iccv/NguyenT19}, FSS-1000~\citep{DBLP:conf/cvpr/LiWCTT20} and PerSeg~\citep{zhang2023personalize} to evaluate the performance of our proposed IPSeg network in the open-world scene. 
Note that PerSeg is a new dataset collected by PerSAM, which comprises a total of 40 objects from various categories, including daily necessities, animals, and buildings. For each object, there are 5 to 7 images and masks, representing different poses or scenes.
We use the same setting in the paper PerSAM to perform experiments.
Unlike previous few-shot works utilizing the image-mask pair as input, 
our method only needs a randomly sampled image as the image prompt.

Moreover, inspired by the work ViL-Seg~\citep{liu2022open}, we employ three datasets, including COCO-Stuff~\citep{DBLP:conf/cvpr/CaesarUF18}, PASCAL-VOC~\citep{DBLP:journals/ijcv/EveringhamGWWZ10} and PASCAL-Context~\citep{DBLP:conf/cvpr/MottaghiCLCLFUY14} to evaluate the performance of our IPSeg network in the zero-shot setting. 
We use the same experimental setting of ViL-Seg to perform the experiments. 
For the above datasets, \textbf{15} classes (frisbee, skateboard, cardboard, carrot, scissors, suitcase, giraffe, cow, road, wall concrete, tree, grass, river, clouds, playingfield) is out of the 183 object categories in COCO-Stuff; \textbf{5} classes (potted plant, sheep, sofa, train, tv-monitor) is 
out of the 20 object categories in PASCAL-VOC; \textbf{4} classes (cow, motorbike, sofa, cat) is out of the 59 object categories in PASCAL-Context.

\subsection{Quantitative Evaluation}
Herein, we do not utilize the ground-truth mask corresponding to the prompt image to select its foreground object for IPSeg.

\subsubsection{Compared to Generalist Models}
{

We select five representative open-world object segmentation methods that employ foundational models in distinct ways: Painter~\citep{DBLP:conf/cvpr/WangWCS023}, SegGPT~\citep{DBLP:conf/iccv/WangZCWS023}, DeLVM~\citep{guo2024dataefficient}, PerSAM~\citep{zhang2023personalize} and Matcher~\citep{DBLP:journals/corr/abs-2303-02153}. Painter, SegGPT and DeLVM are based on a generalized foundation model that is directly trained for various tasks, allowing the use of image-mask pairs for open-world object segmentation. In contrast, PerSAM and Matcher efficiently adapt SAM for open-world object segmentation tasks without the need for additional training. The comparative results are in Table. \ref{generalist}.

As indicated in Table. \ref{generalist}, our proposed method consistently outperforms Painter, DeLVM and PerSAM. This demonstrates the efficacy of our IPSeg network. Specifically, our approach shows significant mIoU performance improvements over PerSAM on the COCO-20$^i$, FSS, and PerSeg datasets, with improvements of 87.0\%, 1.3\%, and 3.6\%, respectively. A noteworthy point is that Painter, DeLVM and PerSAM rely on image-mask pair inputs, which are more stringent and less flexible approaches compared to our method. This observation suggests that the use of a single image as a prompt, as proposed in our method, is a promising avenue for further research. This approach could serve as an alternative or supplement to the traditional image-mask pair prompts, potentially broadening the scope of research in open-world segmentation tasks.

Note that, our proposed IPSeg is designed for the zero-shot open-world segmentation. Therefore, for fair comparison, we also evaluate the performance of Matcher under zero-shot setting. The zero-shot setting means using the unsupervised salient object detection method TSDN~\citep{DBLP:conf/cvpr/Zhou0YLX23} to filter the background of image prompts instead of their corresponding ground truth. As shown in Table. \ref{generalist}, IPSeg can surpass Matcher's performance in the zero-shot setting (Matcher-Z) by a large margin, which further illustrates the validity of our IPSeg.

}

\begin{figure*}[]
    \centering
    \includegraphics[width=\linewidth]{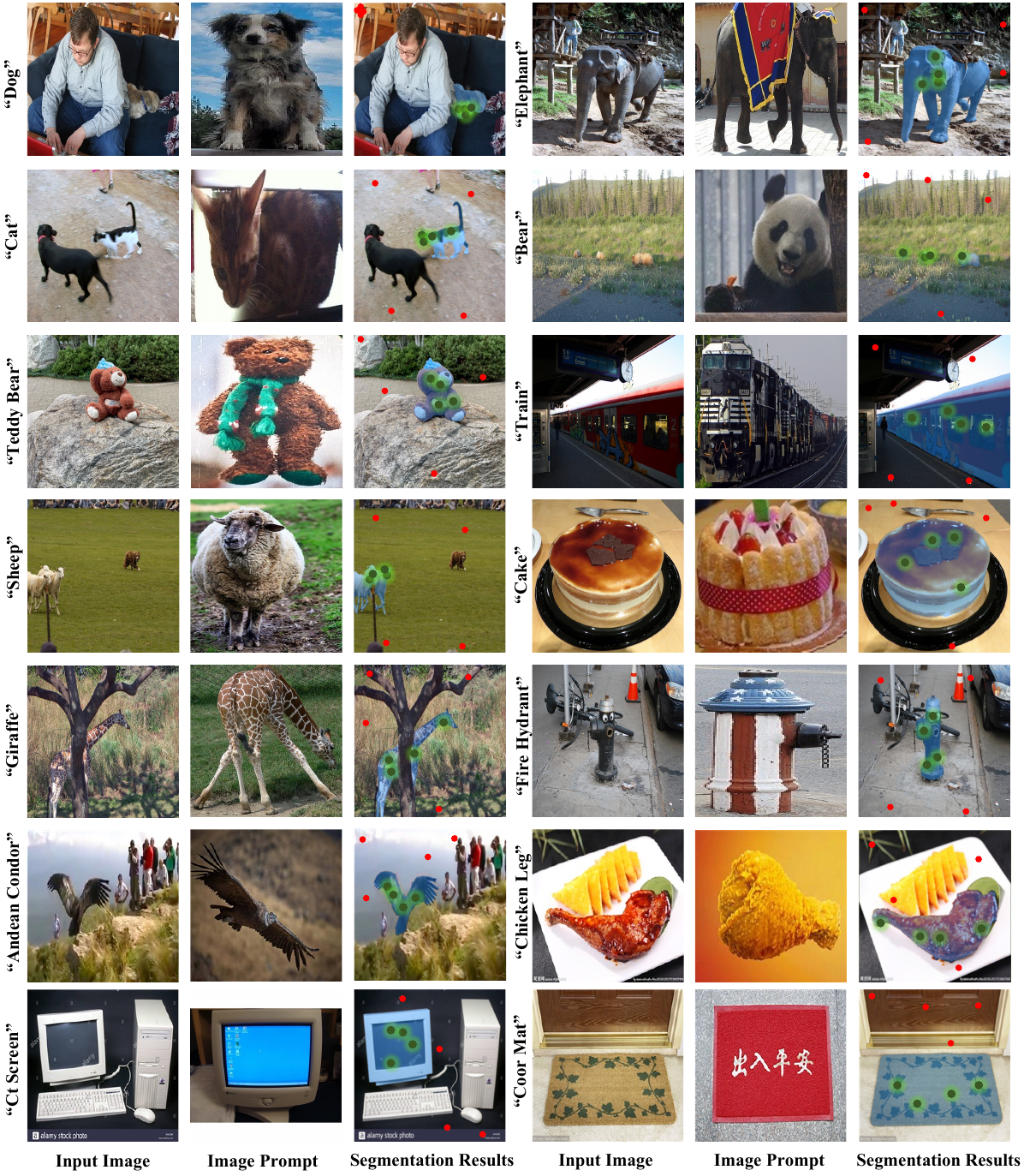}
    \caption{Qualitative segmentation results of the proposed IPSeg framework. It can be seen that the proposed method can effectively segment the objects contained in the prompt image in the input images from different scenarios. 
    The {\color{green} green} point represents positive point prompts sent to SAM, while the {\color{red} red} point represents negative point prompts sent to SAM.}
    \label{Vis}
\end{figure*}

\begin{figure*}[!htbp]
    \centering
    \includegraphics[width=\linewidth]{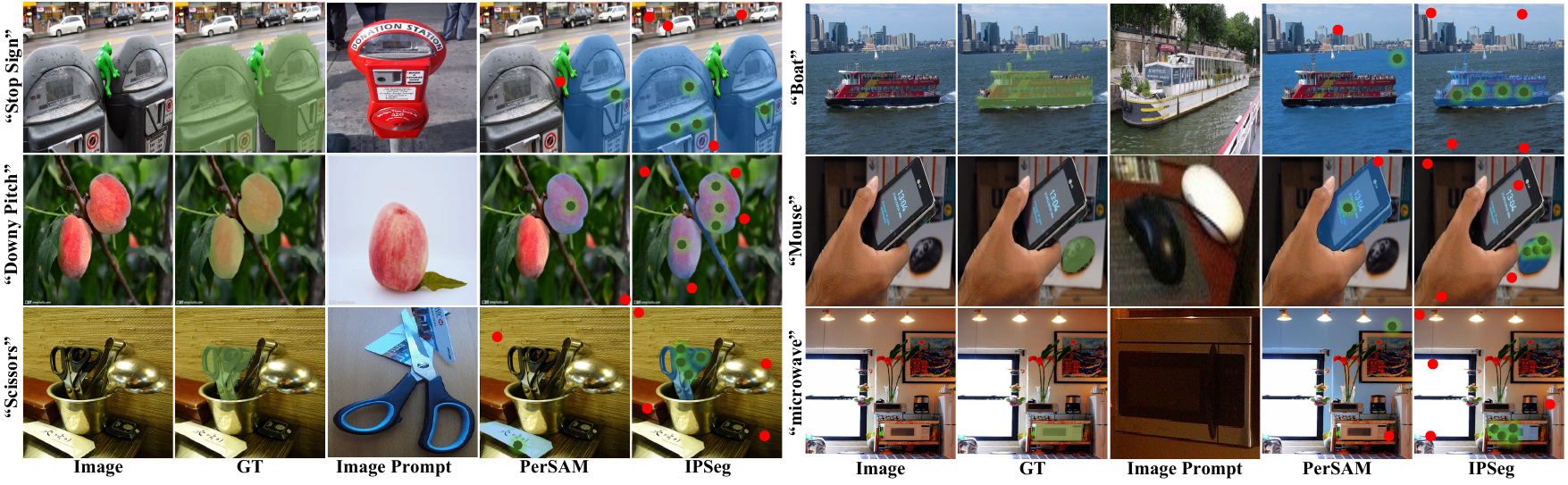}
    \caption{Qualitative segmentation results of the proposed IPSeg and PerSAM using same image prompts. The {\color{green} green} point represents positive point prompts sent to SAM, while the {\color{red} red} point represents negative point prompts sent to SAM.}
    \label{vis_ipseg_persam}
\end{figure*}

\begin{table*}[!htbp]
\centering
\caption{Ablation studies of the combination of SD and DINOv2 in this paper. We report mIoU (\%) in this table. w/o means without.}
\begin{tabular}{@{}c|ccccc|c|c@{}}
\toprule
                  & \multicolumn{5}{c|}{\textbf{COCO-20$^i$}}               &                             &      \\ \cmidrule(lr){2-6}
\multirow{-2}{*}{\textbf{Methods}} &
  \textbf{Fold0} &
  \textbf{Fold1} &
  \textbf{Fold2} &
  \textbf{Fold3} &
  \textbf{Mean} &
  \multirow{-2}{*}{\textbf{FSS}} &
  \multirow{-2}{*}{\textbf{PerSeg}} \\ \midrule
Ours (w/o DINOv2) & {\color[HTML]{333333} 21.6} & 22.0 & 21.8 & 22.2 & 21.9 & {\color[HTML]{333333} 63.5} & 72.9 \\
Ours (w/o SD)     & {\color[HTML]{333333} 39.7} & 43.5 & 39.4 & 44.0 & 41.7 & {\color[HTML]{333333} 80.2} & 90.1 \\
Ours (DINOv2 + SD)               & {\color[HTML]{333333} \textbf{40.9}} & \textbf{44.9} & \textbf{40.1} & \textbf{46.2} & \textbf{43.0} & {\color[HTML]{333333} \textbf{82.7}} & \textbf{92.7} \\ \bottomrule
\end{tabular}
\label{combination}
\end{table*}

\subsubsection{Compared to Specialist Models}
We have conducted a comparison of our proposed IPSeg network with several well-known specialist zero-shot segmentation methods, including SPNet~\citep{xian2019semantic}, ZS3~\citep{bucher2019zero}, CaGNet~\citep{gu2020context}, SIGN~\citep{DBLP:conf/iccv/ChengNNA21}, ViL-Seg~\citep{liu2022open}, GroupVit~\citep{DBLP:conf/cvpr/XuMLBBKW22} and TCL~\citep{DBLP:conf/cvpr/ChaMR23}. It is important to note that these specialist methods are designed with specific segmentation models, each trained on particular datasets. The comparative results are displayed in Table. \ref{Specialist}. Our IPSeg network demonstrates superior performance compared to these specialist models. Notably, it outperforms the CLIP-based ViL-Seg method on the COCO-Stuff, Pascal-VOC, and Pascal-Context datasets, with mIoU performance improvements of 99.4\%, 68.3\%, and a remarkable 231\%, respectively. It is worth mentioning that the Pascal-Context dataset, primarily comprising four common classes, represents relatively simpler scenarios. This aspect may have contributed to the substantial superiority of IPSeg over ViL-Seg in this dataset. Compared to TCL , our method has also achieved competitive performance.

In conclusion, our training-free IPSeg network consistently surpasses specialist open-world object segmentation methods. This success underscores the potential of exploring open-world object segmentation from a novel angle, combining foundational models in a training-free approach. Such an endeavor could significantly enhance the efficiency and applicability of segmentation tasks in diverse real-world scenarios.

\subsection{Qualitative Evaluation}
In Fig. \ref{Vis}, we showcase the visualization results from our IPSeg network. These visualizations highlight the network's capability in effectively segmenting objects within a variety of complex scenes. This serves as a testament to the effectiveness of our approach from a visual standpoint. Particularly noteworthy is the network's performance in intricate scenarios involving multiple objects, such as scenes labelled `Dogs' and `Elephants.' In these cases, our IPSeg network accurately segments the target objects, underscoring its proficiency in correctly identifying objects in the input image that have semantic correspondence with those in the image prompt. This ability showcases the robustness and adaptability of the IPSeg network in dealing with diverse and challenging segmentation tasks.{ To further illustrate the validity of our method, we conduct some visual comparisons with PerSAM in Fig. \ref{vis_ipseg_persam}. It can be seen that in different complex scenes, the performance of our IPSeg is better than that of PerSAM under the same image prompts.}

\begin{figure*}[!htbp]
    \centering
    \includegraphics[width=0.88\linewidth]{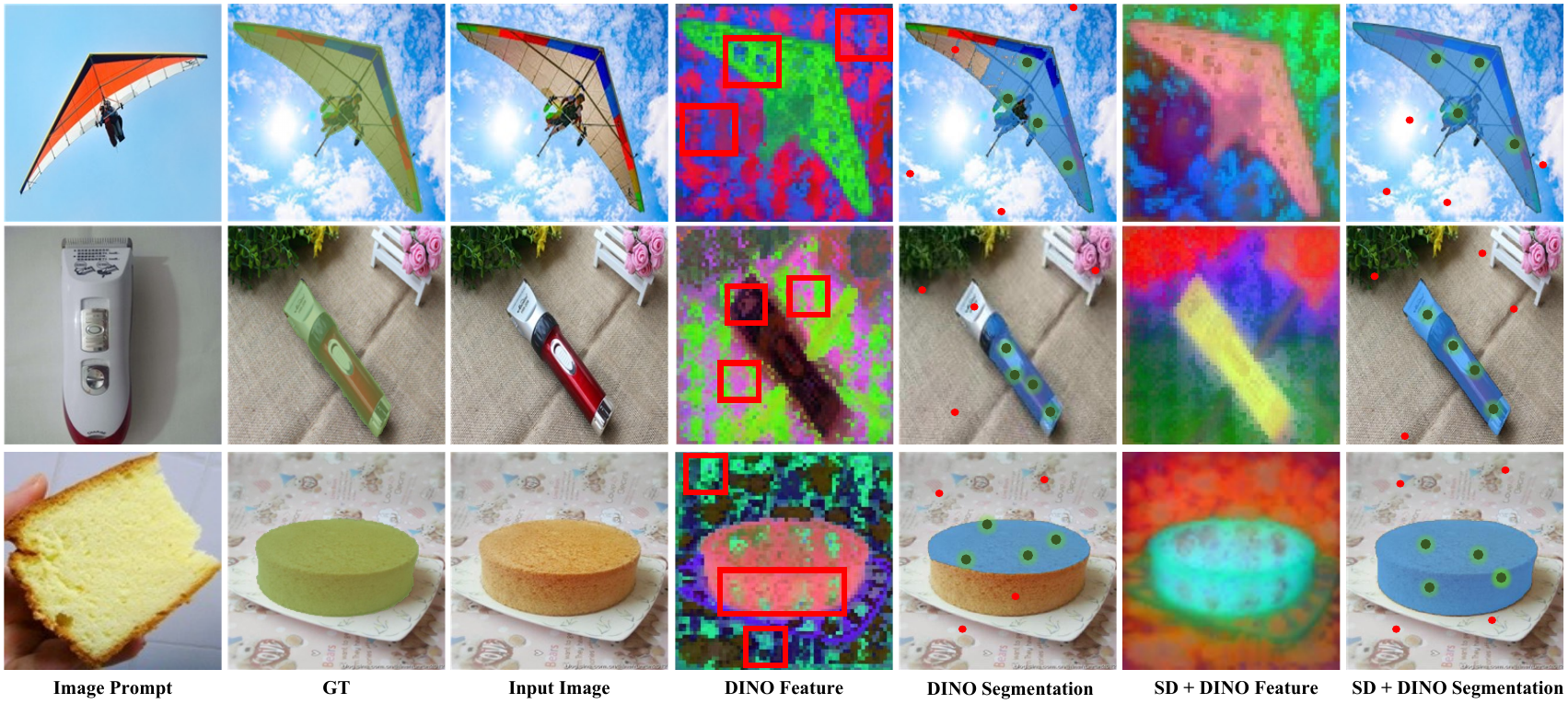}
    \caption{Further analysis about why adding SD can help improve the performance. The second and fifth columns indicate the use of only the DINOv2 model for feature extraction, while the third and sixth columns denote the use of both DINOv2 and SD models for this purpose.}
    \label{vis_r2_q4}
\end{figure*}

\begin{table*}[]
\centering
\caption{Hyperparameters setting in the feature interaction module. We report mIoU (\%) in this table.}
\scalebox{0.95}{
\begin{tabular}{@{}c|ccccc|c|c@{}}
\toprule
\rowcolor[HTML]{FFFFFF} 
\cellcolor[HTML]{FFFFFF}                                   & \multicolumn{5}{c|}{\cellcolor[HTML]{FFFFFF}\textbf{COCO-20$^i$}}                                                                                     & \cellcolor[HTML]{FFFFFF}                               & \cellcolor[HTML]{FFFFFF}                                  \\ \cmidrule(lr){2-6}
\rowcolor[HTML]{FFFFFF} 
\multirow{-2}{*}{\cellcolor[HTML]{FFFFFF}\textbf{Methods}} & \textbf{Fold0}                       & \cellcolor[HTML]{FFFFFF}\textbf{Fold1} & \textbf{Fold2}                       & \textbf{Fold3} & \textbf{Mean} & \multirow{-2}{*}{\cellcolor[HTML]{FFFFFF}\textbf{FSS}} & \multirow{-2}{*}{\cellcolor[HTML]{FFFFFF}\textbf{PerSeg}} \\ \midrule
\rowcolor[HTML]{FFFFFF} 
\cellcolor[HTML]{FFFFFF}Ours(K=32,c=32)                    & 26.1                                 & \cellcolor[HTML]{FFFFFF}27.6           & 26.7                                 & 32.6           & 28.3          & 50.6                                                   & 57.8                                                      \\
\rowcolor[HTML]{FFFFFF} 
Ours(K=32,c=16)                                            & {\color[HTML]{333333} 40.6}          & {\color[HTML]{333333} 44.2}            & {\color[HTML]{333333} 40.4}          & 44.0           & 42.3          & 81.8                                                   & 90.5                                                      \\
\rowcolor[HTML]{FFFFFF} 
Ours(K=32,c=8)                                             & 40.2                                 & 45.0                                   & 40.3                                 & 45.3           & 42.7          & 81.2                                                   & 90.7                                                      \\
\rowcolor[HTML]{FFFFFF} 
\textbf{Ours(K=32,c=4)}                                    & {\color[HTML]{333333} \textbf{40.9}} & {\color[HTML]{333333} \textbf{44.9}}   & {\color[HTML]{333333} \textbf{40.1}} & \textbf{46.2}  & \textbf{43.0} & \textbf{82.7}                                          & \textbf{92.7}                                             \\
\rowcolor[HTML]{FFFFFF} 
Ours(K=32,c=2)                                             & 37.2                                 & 40.5                                   & 37.8                                 & 40.9           & 39.1          & 74.7                                                   & 91.3                                                      \\ \midrule
\rowcolor[HTML]{FFFFFF} 
Ours(K=4,c=4)                                              & 37.5                                 & 40.7                                   & 37.3                                 & 44.7           & 40.1          & 70.0                                                   & 82.2                                                      \\
\rowcolor[HTML]{FFFFFF} 
Ours(K=8,c=4)                                              & 39.0                                 & 43.5                                   & 39.4                                 & 46.2           & 42.0          & 76.1                                                   & 84.1                                                      \\
\rowcolor[HTML]{FFFFFF} 
Ours(K=16,c=4)                                             & 39.9                                 & 44.7                                   & 39.9                                 & 46.3           & 42.7          & 78.2                                                   & 88.3                                                      \\
\rowcolor[HTML]{FFFFFF} 
\textbf{Ours(K=32,c=4)}                                    & {\color[HTML]{333333} \textbf{40.9}} & {\color[HTML]{333333} \textbf{44.9}}   & {\color[HTML]{333333} \textbf{40.1}} & \textbf{46.2}  & \textbf{43.0} & \textbf{82.7}                                          & \textbf{92.7}                                              \\
\rowcolor[HTML]{FFFFFF} 
Ours(K=64,c=4)                                             & 37.8                                 & 42.3                                   & 38.2                                 & 42.0           & 40.1          & 81.6                                                   & 90.9                                                      \\ \bottomrule
\end{tabular}}
\label{Hyperparameters}
\end{table*}

\subsection{Ablation Studies}
For ablation studies, similar to the experimental setting above, we do not utilize the ground truth corresponding to the prompt image to select its foreground object.

\subsubsection{Combination of SD and DINOv2}
In the process of feature extraction, our IPSeg network considers both high-level and low-level details from the input and prompt images. Recognizing the limitations of the DINOv2 model in capturing low-level features, we integrate the SD model to address this gap. As shown in Table. \ref{combination}, incorporating SD significantly boosts the performance of our IPSeg network. This improvement is further evidenced by the visual results in Fig. \ref{dinosd}, where the inclusion of SD is observed to result in smoother feature representations. { Moreover, as shown in Table. \ref{combination}, the performance of using solely SD as the feature extractor is clearly inferior to that of using a combination of DINOv2 and SD. One primary reason is that the features extracted by SD lack high-level semantic information. As illustrated in Fig. \ref{vis_r2_q4}, incorporating features extracted by the SD model allows IPSeg to more distinctly differentiate between foreground and background. This enhancement significantly boosts the performance of IPSeg.}

\begin{table*}[!htbp]
\centering
\caption{Image prompt robustness of this paper. We report mIoU (\%) in this table.}
\scalebox{1.0}{
\begin{tabular}{@{}c|ccccc|c|c@{}}
\toprule
\rowcolor[HTML]{FFFFFF} 
\cellcolor[HTML]{FFFFFF}                                   & \multicolumn{5}{c|}{\cellcolor[HTML]{FFFFFF}\textbf{COCO-20$^i$}}                                                                                     & \cellcolor[HTML]{FFFFFF}                               & \cellcolor[HTML]{FFFFFF}                                  \\ \cmidrule(lr){2-6}
\rowcolor[HTML]{FFFFFF} 
\multirow{-2}{*}{\cellcolor[HTML]{FFFFFF}\textbf{Methods}} & \textbf{Fold0}                       & \cellcolor[HTML]{FFFFFF}\textbf{Fold1} & \textbf{Fold2}                       & \textbf{Fold3} & \textbf{Mean} & \multirow{-2}{*}{\cellcolor[HTML]{FFFFFF}\textbf{FSS}} & \multirow{-2}{*}{\cellcolor[HTML]{FFFFFF}\textbf{PerSeg}} \\ \midrule
\rowcolor[HTML]{FFFFFF} 
Ours(Prompt Set-1)                                         &40.8                                     &42.3                                        &39.4                                      &45.1                &41.9              &82.6                                                       &92.1                                                           \\
\rowcolor[HTML]{FFFFFF} 
Ours(Prompt Set-2)                                         &38.9      &45.9        &40.3      &44.8       &42.5      & 82.5                                            & 91.9                                                 \\
\rowcolor[HTML]{FFFFFF} 
{Ours(Prompt Set-3)}                                & {\color[HTML]{333333} {40.9}} & {\color[HTML]{333333} {44.9}}   & {\color[HTML]{333333} {40.1}} & {46.2}  & {43.0} & {82.7}                                          & {92.7}                                              \\ \bottomrule
\end{tabular}}
\label{robustness}
\end{table*}

\begin{figure*}[]
    \centering
    \includegraphics[width=0.9\linewidth]{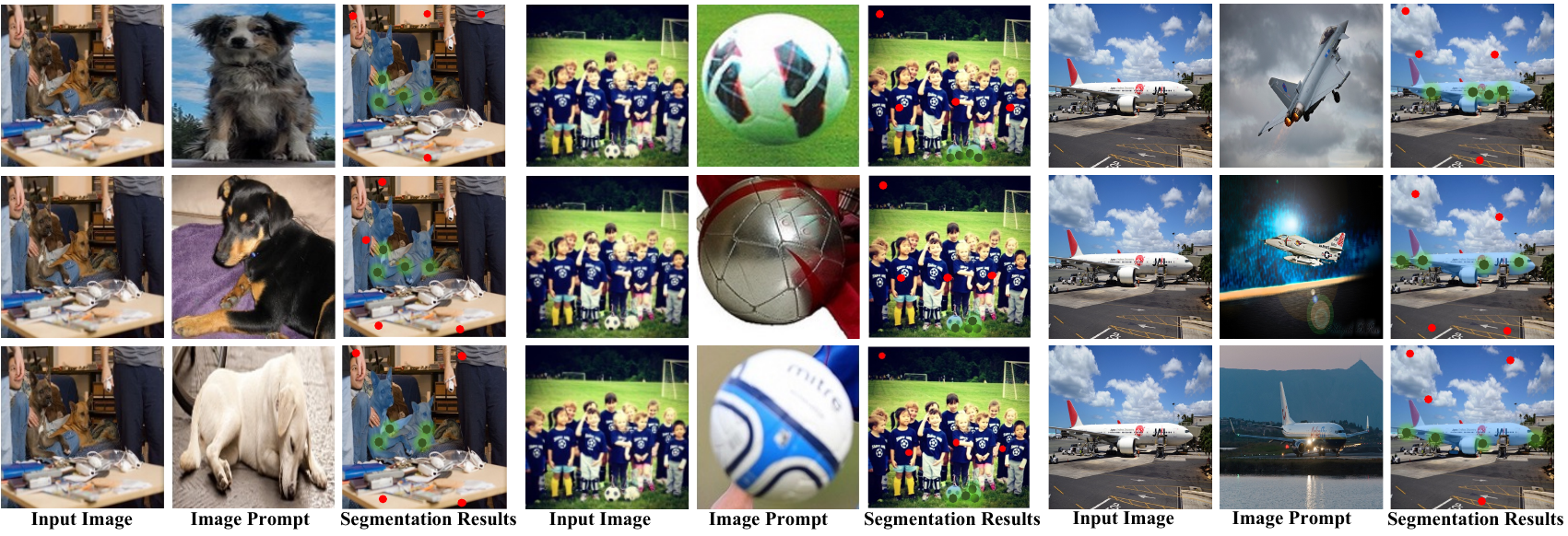}
    \caption{Qualitative results of the proposed IPSeg framework when using different image prompts. When given the same input image with different image prompts, our proposed IPSeg network can consistently generate satisfactory results. This also indicates the robustness of our method. The {\color{green} green} point represents positive point prompts sent to SAM, while the {\color{red} red} point represents negative point prompts sent to SAM.}
    \label{roboust}
\end{figure*}

\begin{figure*}[!htbp]
    \centering
    \includegraphics[width=0.95\linewidth]{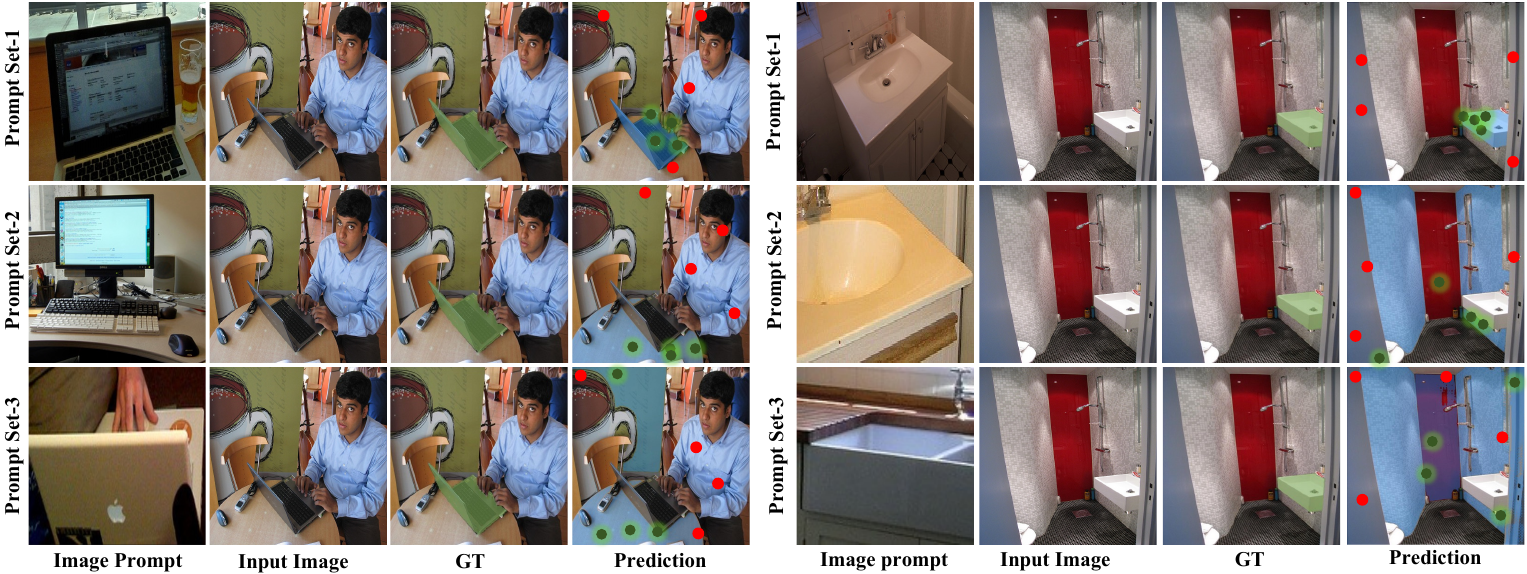}
    \caption{Some failure prediction results of our IPSeg under different image prompts. The {\color{green} green} point represents positive point prompts sent to SAM, while the {\color{red} red} point represents negative point prompts sent to SAM. }
    \label{prompt_diff}
\end{figure*}

\subsubsection{Hyperparameters in Feature Interaction}
In feature interaction, we introduce a simple yet effective approach for generating point prompts to guide SAM in generating the corresponding segmentation results. In this module, we compute the similarity between each pixel in the prompt image and the input image using cosine similarity. We use the TopK algorithm to select the TopK most/least similar points, followed by the clustering algorithm to group these points into $c$ cluster centers. In Table. \ref{Hyperparameters}, we investigate the impact of different values of $K$ and $c$ on performance. We observe that using the TopK algorithm alone helps the model achieve initial performance (Ours($K=32,c=32$)), and further application of the clustering algorithm improves performance even more.

\begin{table*}[]
\centering
\caption{The impact of background noise on IPSeg. We report mIoU (\%) in this table. w/o means without operation}
\scalebox{1.0}{
\begin{tabular}{@{}c|ccccc|c|c@{}}
\toprule
 &
  \multicolumn{5}{c|}{\textbf{COCO-20$^i$}} &
   &
   \\ \cmidrule(lr){2-6}
\multirow{-2}{*}{\textbf{Methods}} &
  \textbf{Fold0} &
  \textbf{Fold1} &
  \textbf{Fold2} &
  \textbf{Fold3} &
  \textbf{Mean} &
  \multirow{-2}{*}{\textbf{FSS}} &
  \multirow{-2}{*}{\textbf{PerSeg}} \\ \midrule
Ours (w/o TSDN) &
  \cellcolor[HTML]{FFFFFF}{\color[HTML]{333333} 8.9} &
  \cellcolor[HTML]{FFFFFF}10.2 &
  \cellcolor[HTML]{FFFFFF}9.1 &
  \cellcolor[HTML]{FFFFFF}9.4 &
  9.4 &
  {\color[HTML]{333333} 32.8} &
  50.2 \\
Ours (with TSDN) &
  {\color[HTML]{333333} 40.9} &
  44.9 &
  40.1 &
  46.2 &
  43.0 &
  {\color[HTML]{333333} 82.7} &
  92.7 \\
Ours (with A2S) &
  {\color[HTML]{333333} 40.1} &
  44.3 &
  39.3 &
  45.9 &
  42.4 &
  {\color[HTML]{333333} 82.5} &
  92.0 \\
Ours (with GT) &
  \cellcolor[HTML]{FFFFFF}44.7 &
  \cellcolor[HTML]{FFFFFF}48.8 &
  \cellcolor[HTML]{FFFFFF}46.9 &
  \cellcolor[HTML]{FFFFFF}48.9 &
  47.3 &
  84.7 &
  93.6 \\ \bottomrule
\end{tabular}}
\label{r2_q3}
\end{table*}

\subsubsection{Image Prompt Robustness}
In this paper, we introduce a more flexible approach to using image prompts. To further validate the robustness of our model with different image prompt combinations, we randomly selected three different image prompt combinations. {  Specifically, we prepare appropriate prompt images based on their categories. For all prompt images, we firstly manually choose different prompt images with clear visual representations in certain classes. Then, we randomly compose prompt set-1 to set-3 from these prompt images. Note that, all prompt images are chosen from the used benchmark based on categories, such as COCO and FSS, and we make sure that the selected prompt image and evaluation datasets do not have the same image.} From Table. \ref{robustness}, it can be observed that our method maintains good robustness across different prompt inputs. As shown in Fig. \ref{roboust}, when given the same input image with different image prompts, our proposed IPSeg network can consistently generate satisfactory results. This experiment further indicates that in future improvement of this framework, researchers can have a more flexible choice of prompts, reaffirming the potential of our IPSeg.

{

Moreover, in Fig. \ref{prompt_diff}, we share some failed cases. Specifically, we showcase objects that are correctly segmented in prompt set-1 but failed in prompt set-2 and set-3. This group of examples indicates that choosing image prompts with only a single, complete target object can significantly aid IPSeg in achieving accurate segmentation results. Hence, when preparing image prompts, we strive to adhere to these two principles for collecting image prompts. However, while aiming for optimal performance, we do not want our framework to be constrained by the reference images. Consequently, the three prompt sets designed in Table. \ref{robustness} are not deliberately combined. This design approach ensures that the results obtained by IPSeg are reliable and credible and indirectly shows that our framework does not rely on carefully selected image prompts that require extensive time investment.}

\begin{table*}[!htbp]
\centering
\caption{The performance of IPSeg using different feature extractors. We report mIoU (\%) in this table.}
\scalebox{0.9}{
\begin{tabular}{@{}c|ccccc|c|c@{}}
\toprule
 & \multicolumn{5}{c|}{\textbf{COCO-20$^i$}} &  &  \\ \cmidrule(lr){2-6}
\multirow{-2}{*}{\textbf{Methods}} &
  \textbf{Fold0} & 
  \textbf{Fold1} & 
  \textbf{Fold2} & 
  \textbf{Fold3} & 
  \textbf{Mean} & 
  \multirow{-2}{*}{\textbf{FSS}} & 
  \multirow{-2}{*}{\textbf{PerSeg}} \\ \midrule 
Ours (MAE) &
  \cellcolor[HTML]{FFFFFF}{\color[HTML]{333333} 9.9} &
  \cellcolor[HTML]{FFFFFF}12.2 &
  \cellcolor[HTML]{FFFFFF}9.9 &
  \cellcolor[HTML]{FFFFFF}11.6 &
  10.7 &
  {\color[HTML]{333333} 36.7} &
  53.7 \\
Ours (CLIP) &
  \cellcolor[HTML]{FFFFFF}{\color[HTML]{333333} 20.3} &
  \cellcolor[HTML]{FFFFFF}26.8 &
  \cellcolor[HTML]{FFFFFF}21.6 &
  \cellcolor[HTML]{FFFFFF}29.9 &
  24.7 &
  {\color[HTML]{333333} 65.0} &
  89.7 \\
Ours (DINOv2) &
  {\color[HTML]{333333} {40.9}} &
  {44.9} &
  {40.1} &
  {46.2} &
  {43.0} &
  {\color[HTML]{333333} {82.7}} &
  {92.7} \\ \bottomrule
\end{tabular}}
\label{review1_q2}
\end{table*}

\begin{table*}[!htbp]
\centering
\caption{Comparison between IPSeg and PerSAM on other four datasets, containing DAVIS2017~\citep{DBLP:journals/corr/Pont-TusetPCASG17}, Pascal-Part~\citep{DBLP:journals/corr/abs-2007-02419}, PACO-Part~\citep{DBLP:conf/cvpr/RamanathanKPWZG23} and LVIS-92$^i$~\citep{DBLP:conf/cvpr/GuptaDG19}. For video object segmentation (VOS), we report $\mathbf{J}$ and $\mathbf{F}$ scores. For part segmentation and semantic segmentation, we report mIoU (\%). } 
\scalebox{0.88}{
\begin{tabular}{@{}c|c|ccccccc@{}}
\toprule
 &
   &
  \multicolumn{2}{c}{\textbf{VOS-DAVIS2017}} &
  \textbf{} &
  \multicolumn{2}{c}{\textbf{Part Segmentation}} &
  \textbf{} &
  \textbf{Semantic Segmentation} \\ \cmidrule(lr){3-4} \cmidrule(lr){6-7} \cmidrule(l){9-9} 
\multirow{-2}{*}{\textbf{Methods}} &
  \multirow{-2}{*}{\textbf{Pub. \& Year}} &
  \cellcolor[HTML]{FFFFFF}\textbf{$\mathbf{J}$} &
  \cellcolor[HTML]{FFFFFF}\textbf{$\mathbf{F}$} &
  \textbf{} &
  \cellcolor[HTML]{FFFFFF}\textbf{Pascal-Part} &
  \textbf{PACO-Part} &
  \textbf{} &
  \textbf{LVIS-92$\mathbf{^i}$} \\ \midrule
PerSAM &
  ICLR 2024 &
  71.3 &
  75.1 &
   &
  32.5 &
  22.5 &
   &
  15.6 \\
\cellcolor[HTML]{FFFFFF}IPSeg &
  Year 2024 &
  \cellcolor[HTML]{FFFFFF}\textbf{75.3} &
  \cellcolor[HTML]{FFFFFF}\textbf{77.3} &
   &
  \cellcolor[HTML]{FFFFFF}\textbf{34.3} &
  \textbf{29.0} &
   &
  \textbf{20.3} \\ \bottomrule
\end{tabular}}
\label{transferability}
\end{table*}

\subsubsection{Impact of background noise} {
To investigate the impact of background noise on our method, we conduct experiments under the following settings: without using the unsupervised salient object detection (USOD) method TSDN~\citep{DBLP:conf/cvpr/Zhou0YLX23} to filter the background, using TSDN to filter the background, and using the ground truth corresponding to the referring image to filter the background. The results are shown in Table. \ref{r2_q3}. Initially, it is evident that not filtering the background significantly affects our experimental performance. Further improvements are observed upon utilizing TSDN. Finally, by utilizing the ground truth to filter the background noise in the referring image, we can achieve best performance. Moreover, if we use another USOD method A2S~\citep{DBLP:journals/tcsv/ZhouCYXL23}, IPSeg's performance does not fluctuate dramatically. This experiment shows that IPSeg requires the USOD method to provide a relatively less noisy image prompt, but is not dependent on a particular USOD method. }

\subsubsection{Different Feature Extractors} 
{ To demonstrate the impact of different feature extractors, inspired by Matcher~\citep{DBLP:journals/corr/abs-2303-02153}, we use MAE~\citep{he2022masked} and CLIP~\citep{DBLP:conf/icml/RadfordKHRGASAM21} as feature extractors, and the performance is shown in Table. \ref{review1_q2}. Using DINOv2 as feature extractor achieves the best performance on all datasets. Additionally, this experiment demonstrates that IPSeg, as a training-free framework, facilitates the integration of various feature extractors.}

\subsubsection{Transferability of IPSeg} {
We conduct experiments on several datasets to further demonstrate the effectiveness and transferability of our IPSeg, as shown in Table. \ref{transferability}. These datasets contain video object segmentation benchmark DAVIS2017~\citep{DBLP:journals/corr/Pont-TusetPCASG17}, semantic segmentation benchmark LVIS-92$^i$~\citep{DBLP:conf/cvpr/GuptaDG19}, and part segmentation benchmarks Pascal-Part~\citep{DBLP:journals/corr/abs-2007-02419} and PACO-Part~\citep{DBLP:conf/cvpr/RamanathanKPWZG23}. As shown in Table. \ref{transferability}, the performance of our method can outperform PerSAM for all datasets. This point once again illustrates the validity of our IPSeg.}

\section{Conclusion}
In this paper, we introduce the IPSeg framework for open-world segmentation using visual concepts from a single image. IPSeg is a simple yet highly effective approach designed to inspire researchers to approach open-world segmentation from two pivotal perspectives: efficient utilization of foundational models and a flexible setup for prompt information. Through our exploration of how to optimally combine diverse foundational models, our method attains outstanding performance on six widely utilized datasets. Furthermore, our research underscores the importance of adaptability in foundational models, emphasizing their potential to revolutionize the way we approach complex computer vision challenges. We believe that our contributions will pave the way for future research endeavors, pushing the boundaries of what's possible in open-world segmentation and setting new standards for efficiency and versatility in the field.

\section{Data Availability Statement}
All the data used in this study are available from third-party institutions. Researchers can access the data through the instructions presented in the original works of the corresponding datasets. However, researchers should follow specific regulations stated by these datasets and use them for only academic purposes. 

\section{Code Availability}
The code of this work is released at \url{https://github.com/luckybird1994/IPSeg}.

\bibliography{sn-bibliography}
\end{document}